\def\year{2022}\relax
\documentclass[letterpaper]{article} 
\usepackage{aaai22}  
\usepackage{times}  
\usepackage{helvet}  
\usepackage{courier}  
\usepackage[hyphens]{url}  
\usepackage{graphicx,epstopdf} 
\usepackage{natbib}  
\usepackage{caption} 
\DeclareCaptionStyle{ruled}{labelfont=normalfont,labelsep=colon,strut=off} 
\usepackage{amsmath}
\usepackage{multirow}
\usepackage[font=footnotesize]{subcaption}
\usepackage{booktabs}
\usepackage{amssymb}
\usepackage{bbding}
\usepackage{epsfig}
\usepackage{bbding}
\usepackage{pifont}

\newcommand{\etal}{\emph{et al. }}
\newcommand{\ie}{\emph{i.e. }}
\newcommand{\eg}{\emph{e.g. }}
\title{Spatio-Temporal Recurrent Networks for Event-Based Optical Flow Estimation}
 \author{
     Ziluo Ding,
     Rui Zhao,
     Jiyuan Zhang,
     Tianxiao Gao,
     Ruiqin Xiong,
     Zhaofei Yu\thanks{Corresponding author.},
     Tiejun Huang
 }
 \affiliations{
     Peking University\\
     \{ziluo, gtx, rqxiong, yuzf12, tjhuang\}@pku.edu.cn, \{ruizhao, jyzhang\}@stu.pku.edu.cn
 }

\begin{document}
\maketitle
\begin{abstract}
Event camera has offered promising alternative for visual perception, especially in high speed and high dynamic range scenes. Recently, many deep learning methods have shown great success in providing promising solutions to many event-based problems, such as optical flow estimation. However, existing deep learning methods did not address the importance of temporal information well from the perspective of architecture design and cannot effectively extract spatio-temporal features. Another line of research that utilizes Spiking Neural Network suffers from training issues for deeper architecture. To address these points, a novel input representation is proposed that captures the events’ temporal distribution for signal enhancement. Moreover, we introduce a spatio-temporal recurrent encoding-decoding neural network architecture for event-based optical flow estimation, which utilizes Convolutional Gated Recurrent Units to extract feature maps from a series of event images. Besides, our architecture allows some traditional frame-based core modules, such as correlation layer and iterative residual refine scheme, to be incorporated. The network is end-to-end trained with self-supervised learning on the Multi-Vehicle Stereo Event Camera dataset. We have shown that it outperforms all the existing state-of-the-art methods by a large margin. The code link is https://github.com/ruizhao26/STE-FlowNet. 
\end{abstract}

\section{Introduction}
We have witnessed the prosperous development of neuromorphic cameras, which offer promising alternatives for visual perception. For example, one of the most popular neuromorphic vision sensors, the event camera \cite{lichtsteiner2008128,6889103}, has exhibited better potential in handling high-speed scenarios and demonstrated superiority in robustness compared with frame-based cameras. Biologically inspired by the retinal periphery, each pixel responds independently to the change of luminance by generating asynchronous events with microsecond accuracy in the event camera. Therefore, event camera greatly reduces the amount of data needed to be processed, and their distinct and novel working principles provide some promising advantages such as extremely low latency, high dynamic range, and low power consumption, making them a good fit for domains such as optical flow, object tracking, or dynamic scene understanding. Optical flow estimation has a wide range of applications, especially in autonomous driving \cite{aufrere2003perception,capito2020optical} and action recognition \cite{efros2003recognizing}. Although optical flow estimation has achieved remarkable success in frame-based vision \cite{teed2020raft,wang2020cot,liu2020learning}, it is unlikely to directly apply traditional frame-based optical flow estimation algorithms to event data since the data formats in the two fields are too different. 

Recently, many research studies focus on training an end-to-end neural network to estimate optical flow in a self or unsupervised manner \cite{zhu2019unsupervised,ye2018unsupervised,lee2020spike,zhu2018ev}. In most existing deep learning works, a series of asynchronous events are first transformed into alternative representations--event image that summarizes events into a 2D grid. Then, all the event images or feature maps are concatenated as one entity and fed into Convolutional Neural Network (ConvNet) at one time. Although this practice contains spatial information about scene edges that is familiar to conventional computer vision, it did not address the importance of temporal information well from the perspective of architectural design since ConvNet is originally designed to extract spatial hierarchies of features \cite{yamashita2018convolutional}. In addition, event cameras bear severe noise \cite{wang2020joint}. Unfired events can be easily introduced especially in some complex textured and high-speed scenarios, \eg driving on the highway, due to bandwidth limitations \cite{hu2021v2e}. Thus, the problem of event signal enhancement has a strong deviation from its image-based counterpart and requires deliberate modifications when applying frame-based models. 




In our work, we address the signal enhancement by proposing a novel input representation that retrieves the missing events in a high-speed period. Our input representation accumulates events based on the temporal distribution of the event stream. Event signal is enhanced when the distribution is more concentrated since high-speed movement will lead to more events in a short moment. In addition, we introduce a spatio-temporal recurrent encoding-decoding neural network architecture (STE-FlowNet) to capture spatio-temporal information effectively. More specifically, the feature encoder has a novel design, which is utilizing Convolutional Gated Recurrent (ConvGRU) \cite{ballas2016delving} to extract feature maps from a series of event images. Note that event images are fed into STE-FlowNet separately by the temporal order, not as one entity. Also, every timestep the recurrent encoder gets input, the decoder would estimate optical flow between the first event image to the current event image which can be used for the next estimation. The event stream has recorded the detailed motion dynamic and provided more intermediate information, intuitively it is more accurate to estimate optical flow based on the estimated optical flow information of the previous timestep. 

More importantly, recurrent architecture can remove the restriction in module design and provide more convenience in the training procedure. In more detail, as a core in most frame-based algorithms, the correlation layer \cite{ilg2017flownet} has been shown to provide important cues for optical flow estimation. But it has been missed in all the previous event-based work since it cannot extract features from one entity input. Unlike previous work \cite{zhu2019unsupervised,ye2018unsupervised,zhu2018ev}, STE-FlowNet processes data frame by frame, allowing the correlation layer to be incorporated to extract extra features. Although event image has an amount of motion blur, it preserves abundant spatial information which is compatible with the correlation layer. Furthermore, inspired by traditional optimization-based approaches, we iteratively estimate the residual flows to refine final results. Besides, multiple intermediate supervised signals are possible to improve the performance of the network since recurrent architecture outputs multiple optical flows corresponding to the different time windows. 

We evaluate our method on the Multi Vehicle Stereo Event Camera dataset (MVSEC) \cite{zhu2018multivehicle} and demonstrate its superior generalization ability in different scenarios. Results show that STE-FlowNet outperforms all the existing state-of-the-art methods \cite{zhu2019unsupervised,lee2020spike}. Especially for \(dt=4\) case, we obtain a significant accuracy boost of 23.0 \% on average over the baselines. In addition, we validate various design choices of STE-FlowNet through extensive ablation studies. 


\section{Related work}
\subsection{Event-based Optical Flow}



More recently, deep learning has been applied to event-based optical flow thanks to the introduction of some large-scale event-based benchmarks. Many early works \cite{moeys2016steering,6981783} utilize simple ConvNet to estimate optical flow only based on small datasets. EV-FlowNet, presented by Zhu \etal \cite{zhu2018ev}, can be regarded as the first deep learning work training on large datasets with an encoder-decoder architecture. The event steam was summarized to compact event image preserving the spatial relationships between events as well as most recent temporal information. As an updated version of EV-FlowNet, an unsupervised framework has been proposed by Zhu \etal \cite{zhu2019unsupervised}. In more detail, its loss function is designed to measure the motion blur in the event image. The more accurate the optical flow is, the less motion blur the event images possess. ECN \cite{ye2018unsupervised} follows the encoder-decoder architecture. However, it uses an evenly-cascaded structure, instead of standard ConvNet, to facilitate training by providing easily-trainable shortcuts. In summary, it is difficult for ConvNet to find out the temporal correlation between event images from one entity, which is the limitation all the works mentioned above cannot ignore.  

SNN, as biologically inspired computational models, can deal with asynchronous computations naturally and exploit spatio-temporal features from events directly. Many researches \cite{paredes2019unsupervised,richter2014bio,orchard2013spiking,lee2020spike,giulioni2016event,haessig2018spiking} consider it as a perfect fit for event-based vision task. Although they are energy-efficient and hardware-friendly, there still exists a gap in performance between SNN and Analog Neural Network (ANN) since it is hard to retain gradients in deeper layers. Hybrid architecture, Spike-FlowNet, proposed by Lee \etal \cite{lee2020spike} seems to be another promising candidate for event-based optical flow. It utilizes SNN as an encoder to extract spatio-temporal features and has ANN as a decoder to enable deeper architecture. Note that this work is a kind of progress for addressing the importance of temporal features compared with ANN methods. However, SNN still limits the capability of the whole architecture as evidenced by that it does not outperform standard ANN methods. 

\subsection{Frame-based Optical Flow}
Frame-based optical flow estimation is a classical task in the computer vision area through the years and has been solved well. FlowNet \cite{dosovitskiy2015flownet} is the first end-to-end neural network for optical flow estimation, and Fischer \etal \cite{dosovitskiy2015flownet} propose a large dataset FlyingChairs to train the network via supervised learning. PWC-Net \cite{sun2018pwc} and Liteflownet \cite{hui2018liteflownet} introduce the pyramid and cost volume to neural networks for optical flow, warping the features in different levels of the pyramid and learning the flow fields coarse-to-fine. IRR-Net \cite{hur2019iterative} takes the output from a previous iteration as an input for the next iteration using a weights-shared backbone network to iteratively refine the residual of the optical flow, which demonstrates an iterative method can increase the motion analysis performance for an optical flow estimation network. RAFT \cite{teed2020raft} constructs decoder in the network using ConvGRU, iteratively decoding the correlation and context information in a fixed resolution, showing the promising capability of ConvGRU to extract a spatio-temporal relationship.



\begin{figure}[t]
\centering
\includegraphics[width=\linewidth]{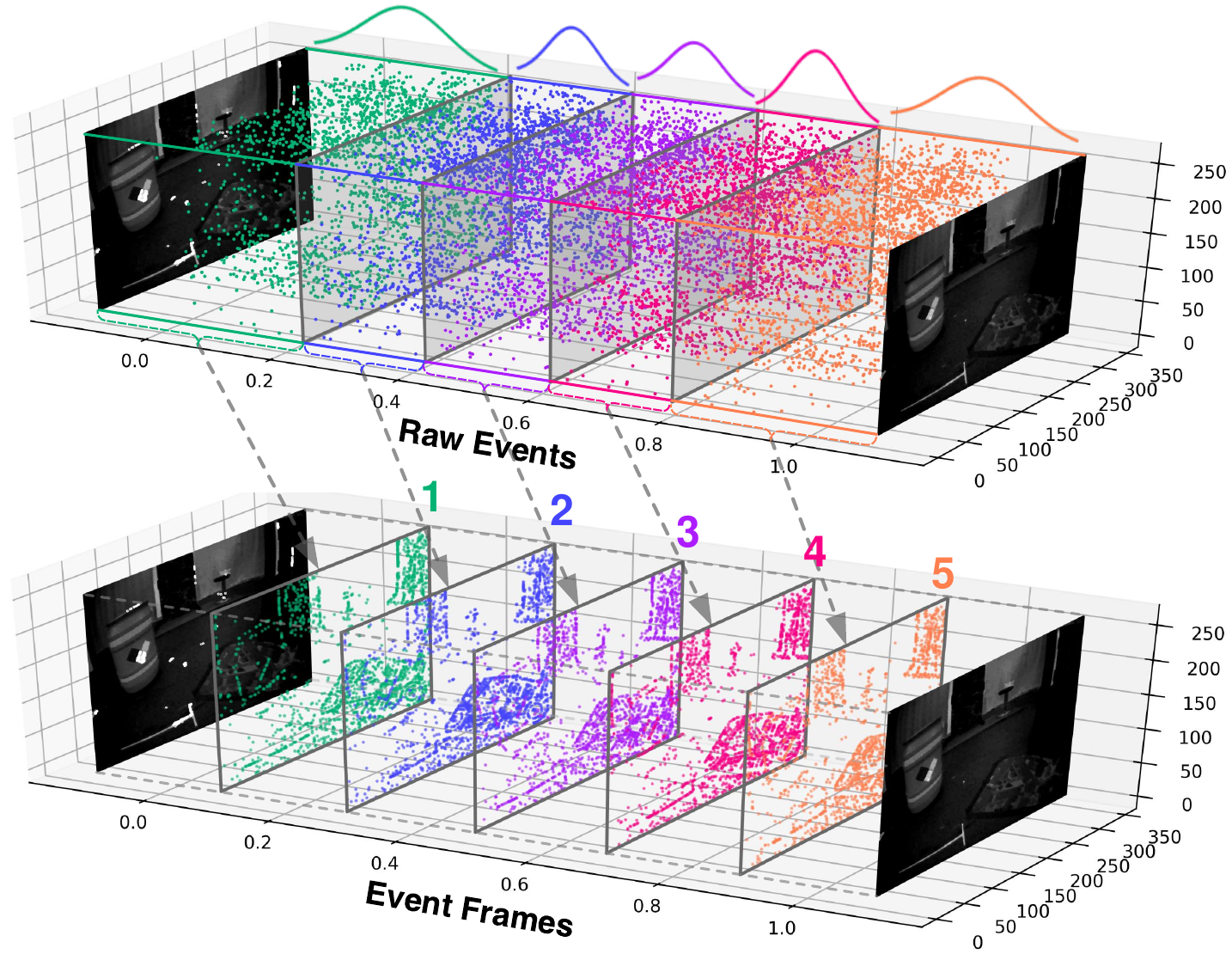}
\caption{Input events representation. (Top) Raw event steam between two consecutive grayscale images from an event camera. (Bottom) Raw event stream are evenly divided to \(N\) splits by number. Each split is converted to a two-channel image, serving as inputs to the network.}
  \label{fig2:input representation}
 \vspace{-0.2cm}
\end{figure}

\section{Method}

Given a series of events from \(t_{\rm{start}}\) to \(t_{\rm{end}}\), we estimate a dense displacement field \( \mathbf{f}=\left( f^1, f^2 \right) \), mapping each pixel \( \mathbf{x}= \left( u, v \right) \) at  \(t_{\rm{start}}\) to its corresponding coordinates  \(  \mathbf{x^{\prime}}=\left( u^{\prime},v^{\prime} \right) = \left( u + f^1(\mathbf{x}),v + f^2(\mathbf{x}) \right)\) at  \(t_{\rm{end}}\).  

\begin{figure}[t]
\centering
\includegraphics[width=0.8\linewidth]{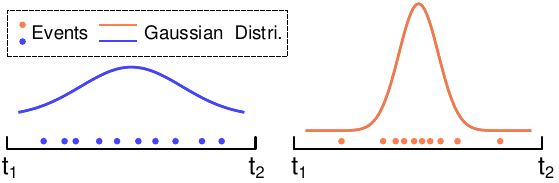}
\caption{Gaussian model is utilized for fitting the events temporal distribution. Summarizing the number of events at each pixel, as well as the last timestamp, cannot capture the temporal distribution of events.}
  \label{fig2:distribution example}
  \vspace{-0.6cm}
\end{figure}

\begin{figure*}[ht]
\centering
\includegraphics[width=\linewidth]{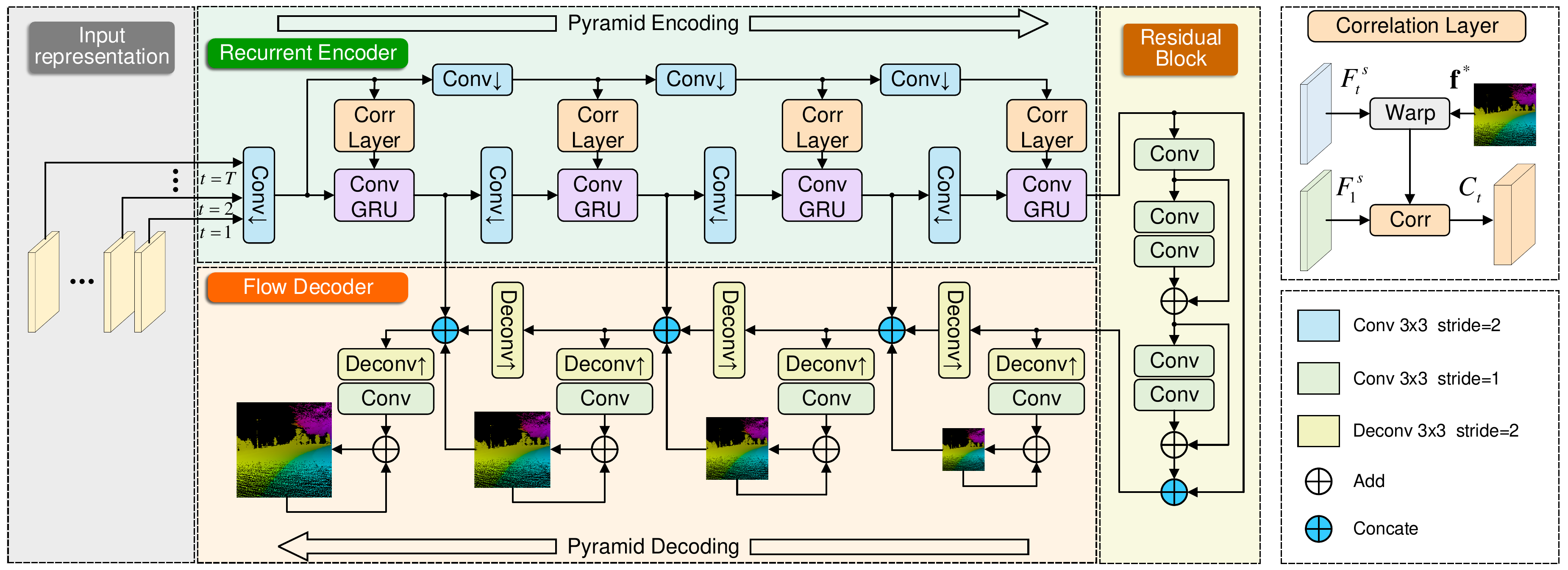}
\caption{Network architecture of the STE-FlowNet. The event input is downsampled through two different paths. One is to get spatial feature maps of four different resolutions for correlation usage later, the other one is through four ConvGRU based encoder modules with skip connections to the corresponding decoders.  After being passed through two residual blocks, the activations are then passed through four decoder modules. In addition, each set of decoder activations is passed through another depthwise convolution layer to generate a flow prediction at its resolution. A loss is applied to this flow prediction, and the prediction is also concatenated to the decoder activations.}
  \label{fig:arch}
  \vspace{-0.4cm}
\end{figure*}

\subsection{Input Data Representation}
Event cameras output a stream of events whenever the log intensity changes exceed a set threshold \(\theta\). Each event encodes the pixel location of brightness change, timestamp of the event and the polarity (increase or decrease of a brightness), which can be summarized as \(e=\{\mathbf{x}, t, p\}\).

However, the output stream might lose some events due to the limitation of pixel bandwidth in some complex textured and high-speed scenarios \cite{hu2021v2e,lichtsteiner2008128}. It is challenging to estimate optical flow in high-speed scenarios, thus we propose a novel input representation for signal enhancement. In more detail, all the events are first evenly divided to \(N\) splits by number. As illustrated in Figure \ref{fig2:distribution example}, the event image by summing the number of events cannot reflect movement details in that period. If most of the events are concentrated in a very short moment, it is more likely that the event camera has experienced a high-speed scene at that moment compared with the smoother distribution. We, hence, incorporated this prior into input representation by accumulating events weighted by the temporal distribution of the events. 

Given a set of \(M\) input events \({(x_i,y_i,t_i,p_i)_{i \in [1:M]}}\) in one split, Gaussian model is utilized for fitting the events temporal distribution and events are weighted accordingly. We generate the event image \(V\), see Figure \ref{fig2:input representation}, as follows:

\begin{align}
\small
& V^{\pm}(x,y)=\sum_{i=1}^M \lceil p_i^{\pm}(x,y)\lambda k_g(t_i) \rceil\\
& k_g(a) = \frac{1}{\sqrt{2\pi} \sigma} \exp \left( - \frac{(a - \mu)^2}{2 \sigma^2} \right)
\end{align}
Where \(k_g(a)\) is a sampling kernel from Gaussian model, \(\mu\) is the mean and \(\sigma\) is the variance. Both are estimated from all timestamps of \(M\) events. \(\lambda\) is normalizing factor and defined as \(\frac{M}{\sum^M_{n=1} k_g(t_n)}\). The more concentrated the temporal distribution, the greater the weights of events. As a result, the event image is enhanced for the possible high-speed moment. Note that all weighted value are rounded up so that events will not be compressed.  



\subsection{Overall Architecture}
The overall architecture of our work, STE-FlowNet, is a variant of the encoder-decoder network \cite{ronneberger2015u} as illustrated in Figure \ref{fig:arch}. Each event image \(\left( \mathbb{R}^{H\times W\times 2} \right)\) is passed into four encoder modules which doubles output channels and downsamples to 1/2 resolution each time. The output feature maps \(\left( \mathbb{R}^{H/8\times W/8\times D} \right)\) from encoder modules then go through two residual blocks. Besides, there is a skip connection that links each encoder to the corresponding decoder module. While passing the decoder module, the input gets upsampled to 2x resolution by transposed convolution. It restores to the original image size after four decoder modules. In addition, STE-FlowNet outputs an intermediate flow at each resolution which is also part of the input for the next decoder. Note that the loss is applied to all the intermediate flows during the training procedure. More details about the parameter settings of architecture can be found in the supplemental material.

\subsection{Feature Extraction }
The key component of feature extraction is ConvGRU. Although ConvGRU plays an important role in some frame-based works \cite{ren2019fusion,teed2020raft}, their ConvGRU modules only focus on processing high-level feature maps from ConvNet top-layers. Note that ConvNet discards local information in their top layers and the temporal variation between images tends to vanish. Therefore, it can hardly capture fine motion information. To address this issue, encoder modules utilize ConvGRU to extract spatio-temporal features from event images at different resolutions to preserve local motion patterns.   

As illustrated in Figure \ref{fig:arch}, the input event image goes through two different pyramid-like architectures at the same time. One way is passing ConvNet to generate spatial feature maps of the current input event image at different resolutions. Another way is through spatio-temporal feature encoders at different resolutions and each encoder module consists of one ConvGRU and one strided convolutional layer. The input of ConvGRU at timestep \(t\) includes three parts, the feature maps \(F_t^{st}\) from the lower-level spatio-temporal feature encoder, the hidden-state of ConvGRU \(h_{t-1}\) from the previous timestep, and the output of the correlation layer \(C_t\) at the same resolution. The ConvGRU is defined by the following equations:
\begin{equation}
\small
 z_t = \sigma ({\rm{Conv}}_{3 \times 3}([h_{t-1}, C_t, F_t^{st}],W_z))
 \end{equation}
 \begin{equation}
 \small
 r_t = \sigma ({\rm{Conv}}_{3 \times 3}([h_{t-1}, C_t, F_t^{st}],W_r))
 \end{equation}
 \begin{equation}
 \small
 \tilde{h}_t={{\rm{tanh}}({\rm{Conv}}_{3 \times 3}[r_t \odot h_{t-1}, C_t, F_t^{st}], W_h)}
 \end{equation}
 \begin{equation}
 \small
 h_t = (1-z_t) \odot h_{t-1} +z_t \odot \tilde{h}_t
 \end{equation}
 


Note that the correlation layer measures the correspondences between the spatial feature maps (output from pyramid-like ConvNet) of the current event image \( F_{t}^s\) and of the first event image \(F_{1}^s\) given a predicted flow \(\mathbf{f}^{\ast}\) from the previous iteration. The correlation layer is defined by the following equation:

\begin{equation}
\small
    C_{t}\left( \mathbf{x} \right) = {\rm{Corr}} \left( F_{t}^s\left( \mathbf{x + f}^{\ast}\right) , F_{1}^s\left( \mathbf{x} \right) \right)
\end{equation}

\begin{table*}[]
    \centering
    \small
    \begin{tabular}{@{}lcccccccc@{}}
    \toprule
          \multirow{2}{*}{\(dt=1\) frame}& \multicolumn{2}{c}{indoor flying1}& \multicolumn{2}{c}{indoor flying2} & \multicolumn{2}{c}{indoor flying3} & \multicolumn{2}{c}{outdoor day1}\\
          \cmidrule{2-3}  \cmidrule{4-5}  \cmidrule{6-7}  \cmidrule{8-9}
         & AEE &\% Outlier &AEE &\% Outlier &AEE &\% Outlier&AEE &\% Outlier\\
         \midrule
         EV-FlowNet \cite{zhu2018ev}  &1.03&2.2&1.72&15.1&1.53  &11.9& 0.49 &0.2\\
         Zhu \etal \cite{zhu2019unsupervised}& 0.58 &\textbf{0.0}&1.02&4.0&0.87&3.0&\textbf{0.32}  &\textbf{0.0}\\
         Spike-FlowNet \cite{lee2020spike} &0.84&\textbf{0.0}&1.28&7.0&1.11  &4.6& 0.47 &\textbf{0.0}\\
         
        Counts+TimeSurface \cite{zhu2018ev}  &0.60& 0.1 &0.81&2.0&0.73  &1.4& 0.45 &\textbf{0.0}\\
        
         Counts \cite{lee2020spike} &0.62& 0.8 & 0.96 &5.5 & 0.87  &3.8& 0.42  &\textbf{0.0}\\

         Ours& \textbf{0.57} &0.1&\textbf{0.79}&\textbf{1.6}&\textbf{0.72}&\textbf{1.3}&0.42  &\textbf{0.0}\\
        \toprule
                    \multirow{2}{*}{\(dt=4\) frame}& \multicolumn{2}{c}{indoor flying1}& \multicolumn{2}{c}{indoor flying2} & \multicolumn{2}{c}{indoor flying3} & \multicolumn{2}{c}{outdoor day1}\\
          \cmidrule{2-3}  \cmidrule{4-5}  \cmidrule{6-7}  \cmidrule{8-9}
         & AEE &\% Outlier &AEE &\% Outlier &AEE &\% Outlier&AEE &\% Outlier\\
         \midrule
    
         EV-FlowNet \cite{zhu2018ev} &2.25&24.7&4.05&45.3&3.45  &39.7& 1.23 &7.3\\
         Zhu \etal \cite{zhu2019unsupervised}&2.18&24.2&3.85&46.8&3.18  &47.8& 1.30 &9.7\\
         Spike-FlowNet \cite{lee2020spike} & 2.24 &23.4&3.83&42.1&3.18&34.8&1.09  &5.6\\
        
  Counts+TimeSurface \cite{zhu2018ev} & 2.02 & 20.1 & 2.85 & 32.1 & 2.49  & 26.9 & 1.14 & 5.5\\
         
         Counts \cite{lee2020spike} & 2.41 &28.2&3.30&39.5&2.86&33.6&1.32  &6.1\\

         Ours& \textbf{1.77} &\textbf{14.7}&\textbf{2.52}&\textbf{26.1}&\textbf{2.23}&\textbf{22.1}&\textbf{0.99}  &\textbf{3.9}\\
         \bottomrule
    \end{tabular}
    \caption{ Quantitative evaluation of our optical flow network compared to EV-FlowNet \cite{zhu2018ev}, Zhu \etal work \cite{zhu2019unsupervised}, Spike-FlowNet \cite{lee2020spike} and other input representations. For each sequence,
Average Endpoint Error (AEE) and \% Outlier are computed. \(dt=1\)
is computed with a time window between two successive grayscale frames, \(dt=4\) is between four grayscale frames.}
    \label{tab:final_results}
   \vspace{-0.3cm}
\end{table*}

The recurrent architecture is naturally set since we process event images frame by frame and ConvGRU is able to extract information from the previous timestep. The reason that we adopt recurrent architecture is that we hypothesize that it is better to estimate optical flow timestep by timestep. It seems more accurate to estimate the flow of \(t_0-t_n\) based on the estimated flow information of \(t_0-t_{n-1}\) (hidden layer of ConvGRU from the previous timestep). 
 
\subsection{Iterative Updates}

\begin{figure}[t]
\centering
\includegraphics[width=\linewidth]{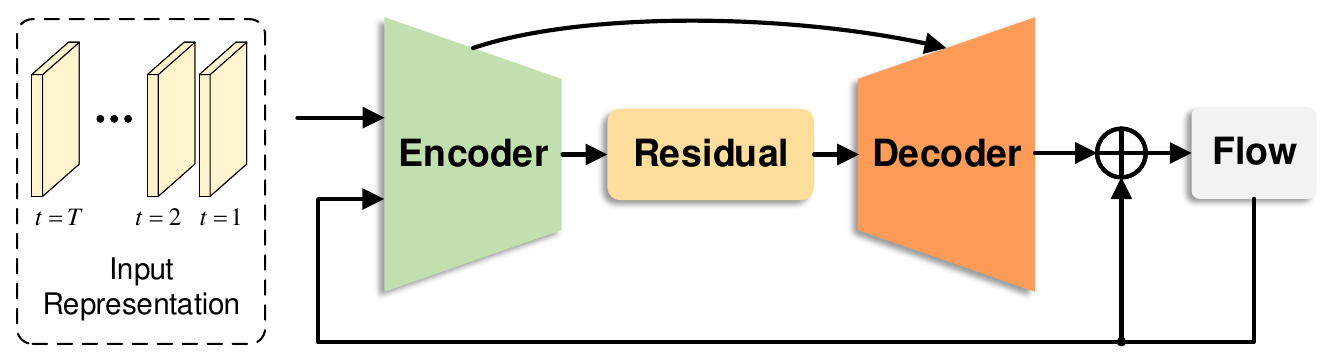}
\caption{Note that next iteration begins after all the event images are passed through the network frame by frame.}
  \label{fig:IRR}
  \vspace{-0.6cm}
\end{figure}

As illustrated in Figure \ref{fig:IRR}, STE-FlowNet adopts the iterative residual refine scheme (IRR) \cite{hur2019iterative} to refine the output, \ie estimated residual flow, iteratively. The final result is the sum of residual
flows from all the iteration steps. Besides, we reuse the same network block with shared weights iteratively. It is noted that the concept of iterative refinement has been proved to be an effective way to improve the final result in many frame-based works \cite{sun2018pwc,hui2018liteflownet,teed2020raft,hur2019iterative}.

Every iteration after the last input event image is fed into STE-FlowNet, we can get a sequence of output \(\{\mathbf{f}_{1,1}^1,\cdots,\mathbf{f}_{T, K}^L\}\), where \(\mathbf{f}_{t,k}^l\) represents the predicted flow between the first event image and \(t^{th}\)  event image at resolution level \(l\) after \(k\) iterations. The new iteration then begins and the event images, as well as the predicted flows, are passed into STE-FlowNet again to refine the flow. More precisely, the predicted flows from the previous iteration will be sent to two places for two different purposes. One is to regard predicted flows as part of the input for STE-FlowNet, which is to warp the current feature maps for correlation use. Another is to perform as a skip connection that adds the predicted flow with the current output of STE-FlowNet to predict the flow at the current iteration. To illustrate, at \(k\) iteration, STE-FlowNet outputs the residual flow \(\Delta{\mathbf{f}_k}\) and the current estimation is \(\mathbf{f}_{k}=\mathbf{f}_{k-1}+\Delta{\mathbf{f}_k}\)  with initial starting point \(\mathbf{f}_{0}=\mathbf{0}\). 

\subsection{Self-Supervised Learning}
Many benchmarks provide synchronized events and grayscale images using DAVIS camera \cite{6889103} so that we can adopt self-supervised learning utilizing grayscale images to guide neural network training. In more detail, the events stream occurring just between two consecutive grayscale images \((I_{t}\), \(I_{t+dt})\) is transformed to multiple event images that are passed into the network subsequently. In the meantime, we can apply the predicted per-pixel flow \(\mathbf{f}=(f^1, f^2 )\) to corresponding grayscale images and generate a self-supervised loss.     

The total loss function consists of two parts \cite{zhu2018ev}, photometric reconstruction loss \(\mathcal{L}_{\rm photo}\) and smoothness loss \(\mathcal{L}_{\rm smooth}\), which can be written as,
\begin{equation}
\small
\mathcal{L}_{\rm total} = \mathcal{L}_{\rm photo}+\lambda\mathcal{L}_{\rm smooth}
\end{equation}
where \(\lambda\) is the weight factor.

The predicted flow is used to warp the second grayscale image using bilinear sampling. The more accurate the predicted flow, the less discrepancy between the first grayscale image and the warped second grayscale image. The photometric loss is computed as follows:
\begin{equation}
\small
\begin{split}
\mathcal{L}_{\rm photo}\left( \mathbf{f}; I_{t}, I_{t+dt} \right) = \sum_{\mathbf{x}} \rho \left( I_{t}\left( \mathbf{x} \right) - I_{t+dt}\left( \mathbf{x} + \mathbf{f} \left( \mathbf{x} \right) \right) \right)
\end{split}
\end{equation}
where \(\rho\) is the Charbonnier loss \(\rho(x)=(x^2+\eta^2)^r\), and we set \(r=0.45\) and \(\eta= 1\mathrm{e}{-3}\).

Furthermore, we use smoothness loss to regularize the predicted flow. It minimizes the flow difference between neighboring pixels, thus it can enhance the spatial consistency of neighboring flows and mitigate some other issues, such as the aperture problem. It can be written as:
\begin{equation}
\small
\begin{split}
\mathcal{L}_{\rm smooth}\left( f^1, f^2 \right) = \frac{1}{HW} \sum_{\mathbf{x}} \vert \nabla  f^1 (\mathbf{x}) \vert + \vert \nabla f^2 (\mathbf{x}) \vert
\end{split}
\end{equation}
where \( \nabla \) is the difference operator, \(H\) is the height and \(D\) is the width of the predicted flow output.

\section{Experiments}
\subsection{Datasets and Training Details}
\label{section:training}
\begin{figure*}[t]
\small
\centering
\includegraphics[width=\linewidth]{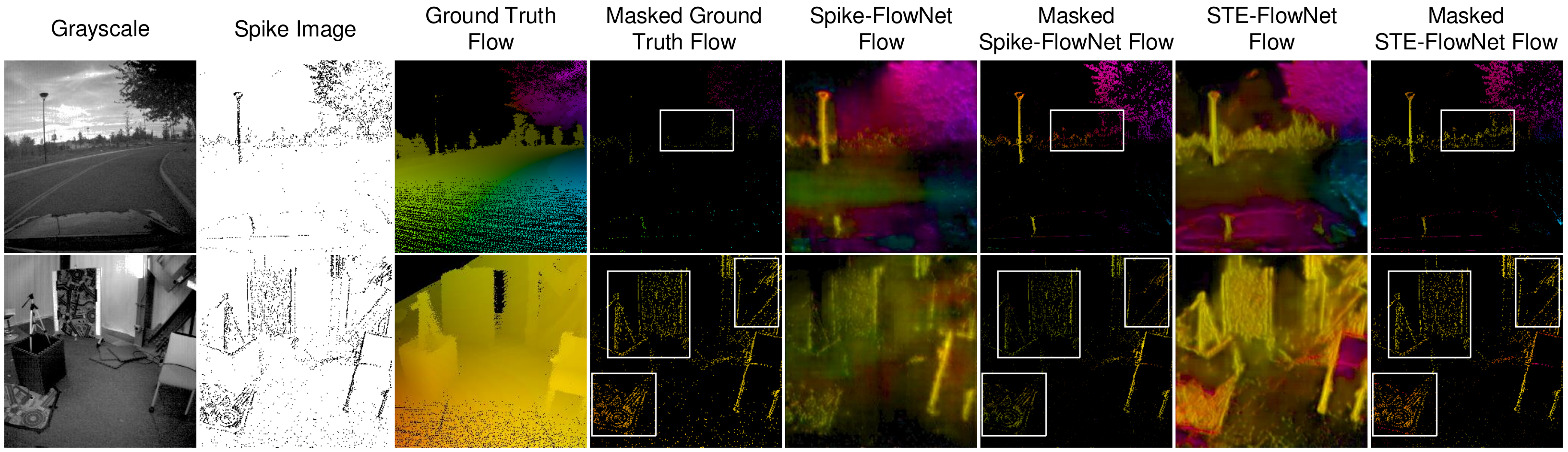}

\caption{Qualitative results from evaluation for \(dt=1\) case. Examples were collected from (Top) outdoor day1 and (Bottom) indoor flying2. The white boxes are used to highlight the superiority of our method.}
\vspace{-0.3cm}
  \label{fig:Qualitative_dt1}
\end{figure*}

The MVSEC dataset \cite{zhu2018multivehicle} is used for training and evaluating our model since it is designed for the development of visual perception algorithms for event-based cameras. Note that the data is collected in two different scenarios, \eg indoor (recorded in rooms) and outdoor (recorded while driving on public roads). Apart from event data sequences and corresponding ground truth optical flow, it also provides the images from a standard stereo frame-based camera pair for comparison which we can use to generate the self-supervised loss. To provide fair comparisons with prior works \cite{zhu2019unsupervised, lee2020spike,zhu2018ev}, we only use the outdoor day2 sequence to train the models. Indoor flying1, indoor flying2, indoor flying3, and outdoor day1 sequences are for evaluation only. 


We have two models for two different time window lengths. In more detail, one is for \(1\) grayscale image frame apart (\(dt=1\)), and the other is for \(4\) grayscale images frame apart (\(dt=4\)). When it comes to \(dt=1\) case, the model is trained for \(40\) epochs. The number of event images \(N_{\rm frame}\) that summarizes input event sequence is set to \(5\), and weight factor \(\lambda\) for the smoothness loss is set to \(10\). The initial learning rate is \(4\mathrm{e}{-4}\) and scaled by \(0.7\) after \(5\), \(10\), and \(20\) epochs. As for \(dt=4\) case, the model is trained for \(15\) epochs. \(N_{\rm frame}\) is set to \(20\) and \(\lambda\) is set to \(10\). In addition, the initial learning rate is \(4\mathrm{e}{-4}\) with the same scaled strategy. Note that we use Adam optimizer \cite{kingma2014adam} with mini-batch size of \(16\) and the number of iteration for IRR \(N_{\rm{irr}}\) is set to \(3\) in both cases. 

As mentioned before, STE-FlowNet would output intermediate flows at different resolutions and the loss computed by each intermediate flow is weighted equally in the final loss. In addition, in the \(dt=4\) cases, there are some more grayscale images available between \({\rm t_{start}}\) and \({\rm t_{end}}\), which means more intermediate self-supervised loss can be generated since recurrent architecture allows STE-FlowNet to predict flows for different time window lengths. The intermediate loss can be used to address vanishing gradients issues and improve the performance of the network \cite{newell2016stacked}. The multiple intermediate losses (MIL) from different time window lengths are then combined to form the final loss with equal weights. 

 \subsection{Quantitative and Qualitative Results }

Average End-point Error (AEE) is used as evaluation metric. It measures the mean distance between the predicted flow \( \mathbf{f}_{\rm{pred}}\) and the ground truth \( \mathbf{f}_{\rm{gt}}\) provided by the MVSEC dataset. Note that only active pixels are reported and they are defined as places where both the ground truth data and the events are present (at least one event can be observed). Besides, we also report the percentage of points with AEE greater than 3 pixels and 5\% of the magnitude of the flow vector, denoted as \% Outlier. During the evaluation, we estimate the optical flow on the center cropped \(256 \times 256 \) of input event images. As for evaluation sequences, we use all the data from the indoor flying sequences. However, only \(800\) grayscale images from the outdoor day1 sequence are chosen, following the settings in \cite{zhu2019unsupervised}. 

\subsubsection{Comparison for Networks}

We compare STE-FlowNet with three existing methods on event-based optical flow estimation:  EV-FlowNet \cite{zhu2018ev}, Spike-FlowNet \cite{lee2020spike} , and the unsupervised framework of  Zhu \etal \cite{zhu2019unsupervised}. Table \ref{tab:final_results} provides the evaluation results in comparison with all the baselines mentioned above. 

For \(dt=1\) case, we can find out that STE-FlowNet has achieved better results compared with all other baselines on three indoor sequences. Specifically, STE-FlowNet achieves AEE of \(0.57\), \(0.79\) and \(0.72\) on the indoor flying1, indoor flying2 and indoor flying3 sequences respectively, \(2\%\), \(23\%\) and \(17\%\) error reduction from the best prior deep network. Note that STE-FlowNet achieves the lowest \% Outlier in most of the evaluation sequences. Although our model doesn't get the best result on outdoor day1,  we have demonstrated the better generalization ability of the model on indoor sequences than baselines. 

\begin{table*}[h]
    \centering
    \small
    \begin{tabular}{@{}lccccccccccccccc@{}}
    \toprule
 & & \multirow{2}{*}{MIL}&\multirow{2}{*}{Corr} & \multirow{2}{*}{IRR}& \multirow{2}{*}{CGRU}& \multicolumn{2}{c}{indoor flying1}&             \multicolumn{2}{c}{indoor flying2}&\multicolumn{2}{c}{indoor flying3}&\multicolumn{2}{c}{outdoor day1}\\
     \cmidrule{7-14} 
        & & & & & & AEE &\% Outlier &AEE &\% Outlier &AEE &\% Outlier&AEE &\% Outlier\\
         \midrule
\multirow{4}{*}{dt=1} &STE-G&\textbf{--}&\ding{55} &\ding{55}&\ding{55}   &1.37 &9.4 &2.16 & 22.7 &2.03 &20.8 &0.59 &0.1\\
&STE-I&\textbf{--}&\ding{55} &\ding{55}&\ding{51}&0.63 &0.2 &0.87 &2.8 &0.79 &2.4 &0.42 &\textbf{0.0} \\
&STE-C&\textbf{--}&\ding{55} &\ding{51} &\ding{51}&0.59 &0.2 &0.87 &2.8 &0.79 & 2.4 &\textbf{0.41 }&\textbf{0.0} \\

&STE&\textbf{--}&\ding{51}  &\ding{51} &\ding{51}  & \textbf{0.57} &\textbf{0.1}&\textbf{0.79}&\textbf{1.6}&\textbf{0.72}&\textbf{1.3}&0.42  &\textbf{0.0}\\
 \midrule
    \midrule
\multirow{5}{*}{dt=4} & STE-G &\ding{51}&\ding{55} &\ding{55}&\ding{55}   &4.77 &65.6 &7.36 &75.8 &6.55 &74.8 &3.10 & 44.5\\
&STE-I&\ding{51}&\ding{55} &\ding{55}&\ding{51} &1.91 &17.7 &2.82 &31.2 &2.41 &24.5 &1.10 & 4.6 \\
&STE-C&\ding{51}&\ding{55} &\ding{51} &\ding{51} &2.11 & 21.9 &2.73 &33.7 &2.46 &27.4 &1.06 &4.1\\
&STE-L&\ding{55}&\ding{51}  &\ding{51} &\ding{51} &1.96 &18.6 &2.70 &30.6 &2.40 &25.7 &1.01 &4.5\\
&STE&\ding{51}&\ding{51}  &\ding{51} &\ding{51} & \textbf{1.77} &\textbf{14.7}&\textbf{2.52}&\textbf{26.1}&\textbf{2.23}&\textbf{22.1}&\textbf{0.99}  &\textbf{3.9}\\
\bottomrule
    \end{tabular}
    \caption{Ablation studies of our design choices for \(dt=1\) and \(dt=4\) case. STE-FlowNet without IRR, and STE-FlowNet without ConvGRU and IRR. The ablation baselines are denoted as STE-C, STE-I, and STE-G respectively. For \(dt=4\), STE-I means removing multiple intermediate losses. }
    \label{tab:alabtion_dt1}
    \vspace{-0.6cm}
\end{table*}

As for \(dt=4\) case, our model has made a remarkable achievement. We outperform all existing approaches in all the test sequences by a large margin. In more detail, we get AEE of \(1.77\), \(2.52\), and \(2.23\) on the indoor flying1, indoor flying2, and indoor flying3 sequences respectively. The error reduction from the best prior work on indoor sequences is \(18\%\), \(34\%\), and \(30\%\). Even on the outdoor day1 sequence, we still have a satisfying result. Our model achieves AEE of \(0.99\) and gets \(9\%\) error reduction. Furthermore, STE-FlowNet achieves the lowest \% Outlier in all the evaluation sequences. The results show that recurrent architecture can better handle more input data since it processes data frame by frame, especially in the scenario where the time window length is long and the number of input event images is large. On the contrary, standard ConvNet architecture might be overwhelmed by massive data at one time.

 The grayscale, spike event, ground truth flow, and the corresponding predicted flow images are visualized in Figure \ref{fig:Qualitative_dt1} where the images are taken from outdoor day1, indoor flying2, and indoor flying3 in \(dt=1\). Since the event data is quite sparse, STE-FlowNet doesn't predict flows in most of the regions. In summary, the results show that STE-FlowNet can accurately estimate the optical flow in both indoor and outdoor scenes. In addition, STE-FlowNet can estimate flows in more edge regions where Spike-FlowNet has no output. Also, in some regions with rich texture, the directions of predicted flows (viewed in color) of STE-FlowNet are closer to the ground truth than Spike-FlowNet.

\subsubsection{Comparison for Input Representations}
We compare our input representation with that of other optical flow algorithms. In more detail, Spike-FlowNet \cite{lee2020spike} generates an event image by summing the number of events at each pixel, denoted as Counts. Ev-FlowNet \cite{zhu2018ev} proposes input
representation that summarizes the number of events at each pixel as well as the last timestamp, denoted as Counts+TimeSurface. For a fair comparison, the backbones of Counts and Counts+TimeSurface are the same as that of STE-FlowNet.


From Table \ref{tab:final_results}, our representation has demonstrated superiority over other methods. Counting events \cite{lee2020spike} discards the rich temporal information in the events, and is susceptible to motion blur. The time Surface method used in \cite{zhu2018ev} is only able to capture some temporal information around some pre-specific moment. Different from these representations, our method accumulates the events based on the temporal distribution of the events stream. We aim to enhance the signal and highlight the period when the event camera encounters a high-speed scene.  

\subsubsection{Qualitative Results}

The grayscale, spike event, ground truth flow, and the corresponding predicted flow images are visualized in Figure \ref{fig:Qualitative_dt1} where the images are taken from outdoor day1, indoor flying2, and indoor flying3 in \(dt=1\). Since the event data is quite sparse, STE-FlowNet doesn't predict flows in most of the regions. In summary, the results show that STE-FlowNet can estimate flows in more edge regions where Spike-FlowNet has no output. Also, in some regions with rich texture, the directions of predicted flows (viewed in color) of STE-FlowNet are closer to the ground truth than Spike-FlowNet. 

\begin{figure}[t]
\small
\vspace{0.2cm}
\centering
\includegraphics[width=\linewidth]{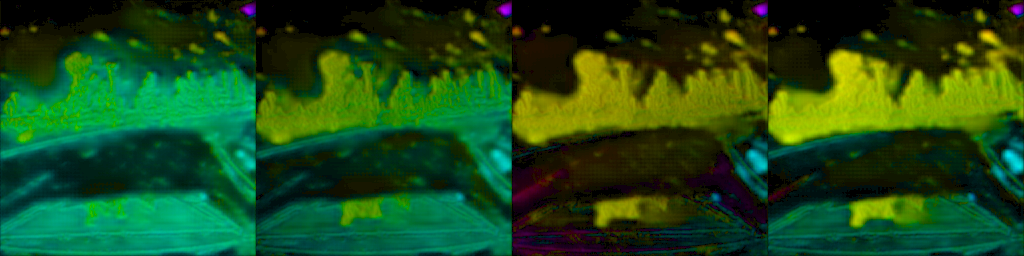}\\
\includegraphics[width=\linewidth]{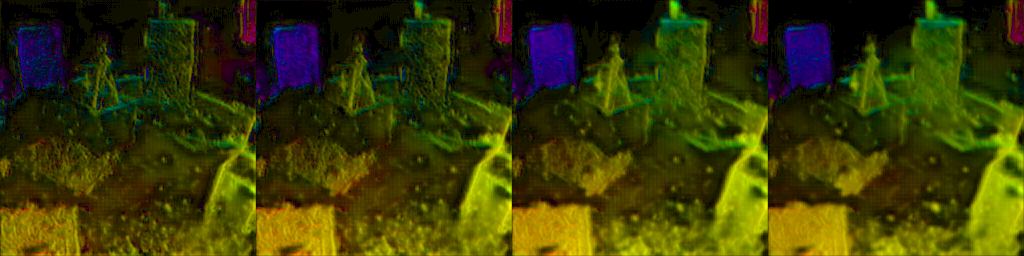}\\
\includegraphics[width=\linewidth]{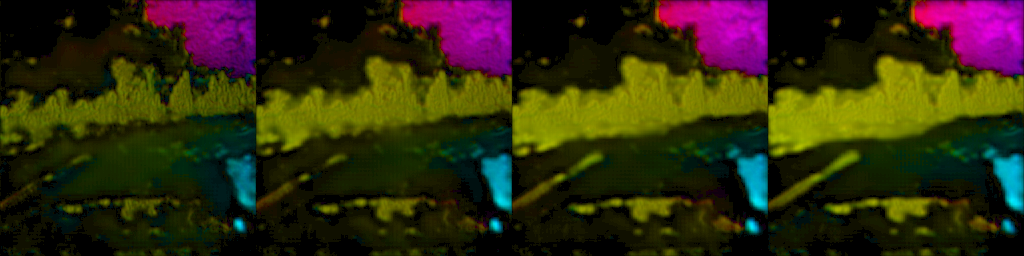}

  \caption{The intermediate outputs from parts timesteps. Timestep increases from left to right.}
\vspace{-0.6cm}
  \label{fig:iterative results}
  
\end{figure}

 Moreover, Figure \ref{fig:iterative results} shows the intermediate outputs from parts timestep of the network. We can find out the optical flow is gradually enhanced and the network indeed can get optical flow information from previous timesteps.

\subsection{Ablation Studies}
 
 There are three components needing to be assessed, \ie ConvGRU, correlation layer, and IRR scheme. Therefore, we have three baselines for ablation studies, namely, STE-FlowNet without correlation layer, STE-FlowNet without IRR, and STE-FlowNet without ConvGRU and IRR. The ablation baselines are denoted as STE-C, STE-I, and STE-G respectively. Note that the correlation layer is meaningless when the IRR scheme is absent since the flows used for correlation come from the previous iteration. Besides, we concatenate the predicted flow with the input event image directly, serving as input for the next iteration in STE-C. As for \(dt=4\) case, we have additional ablation for removing multiple intermediate losses (MIL) for different time windows, denoted as STE-L. 
 
 Table \ref{tab:alabtion_dt1} shows the results of ablation studies in terms of both AEE and \% Outliers for \(dt=1\) case. STE-G performs the worst compared with others in all evaluation sequences. This is because it only depends on one input event image to predict flows without the ConvGRU to provide extra spatio-temporal information from other event images. STE-FlowNet and STE-C perform better than STE-I in almost all evaluation sequences, which demonstrates the effectiveness of the IRR scheme. In addition, STE-FlowNet outperforms STE-C. It shows correlation layer is able to provide more valuable features than directly sending flows. Also, it proves that the correlation layer can be applied in dealing with the event data. Note that STE-I can still achieve promising results compared with some prior works. It again demonstrates the superiority of our architecture. The same conclusions can be obtained in \(dt=4\) case. Moreover, we find out that STE-FlowNet performs better than STE-L. Therefore, we believe that the multiple intermediate losses from different time windows indeed help to improve the performance of the model. Note that the prior works \cite{zhu2019unsupervised,zhu2018ev,ye2018unsupervised,lee2020spike} are unable to utilize additional grayscale images.

\section{Conclusion}
     We propose a ConvGRU-based encoding-decoding network with a novel input representation to effectively extract the spatio-temporal information from event input. Moreover, the correlation layer is used to provide more valuable clues for the IRR scheme to further refine the predicted flow. Empirically, results show that STE-FlowNet outperforms all existing methods in a variety of indoor and outdoor scenes.

\section*{Acknowledgements}
This work was supported by the National Natural Science Foundation of China (62176003, 62088102, 61961130392), Beijing Major Science and Technology Project (Z191100010618003), PKU-Baidu Fund (Project 2019BD001).
\bibliography{aaai22}

\end{document}